%% file: main.tex
\documentclass[a4paper,11pt]{article}
\usepackage[margin=1in]{geometry} 
\usepackage[T1]{fontenc}
\usepackage{graphicx}
\usepackage{graphicx}
\usepackage{xcolor}
\usepackage{multirow}
\usepackage{array}
\usepackage{amsmath} 
\usepackage{algorithm2e}
\usepackage{amssymb}
\usepackage{booktabs}
\usepackage{capt-of}
\usepackage{authblk}
\usepackage{url}


\begin{document}
%
\title{Improving Temporal Action Segmentation via Constraint-Aware Decoding}
%
%
\author[1,2]{Yeo Keat Ee}
\author[3] {Debaditya Roy}
\author[1,2]{Chen Li}
\author[1,2]{Hao Zhang}
\author[1,2,4]{Basura Fernando}

%

%

\affil[1]{Institute of High-Performance Computing, Agency for Science, Technology and Research, Singapore}
\affil[2]{Centre for Frontier AI Research, Agency for Science, Technology and Research, Singapore}
\affil[3]{Indian Institute of Technology Kharagpur, India}
\affil[4]{College of Computing and Data Science, Nanyang Technological University, Singapore}

\date{}

\maketitle              
\begin{abstract}
Temporal action segmentation (TAS) divides untrimmed videos into labeled action segments. While fully supervised methods have advanced the field, challenges such as action variability, ambiguous boundaries, and high annotation costs remain, especially in new or low-resource domains.
Grammar-based approaches improve segmentation with structural priors but rely on complex parsing limiting scalability.
In this work, we propose a lightweight, constraint-based refinement framework that enhances TAS predictions by integrating statistical structural priors such as transition confidence, action boundary sets, and per-class duration, that can be directly extracted from annotated data. These constraints are integrated into a modified Viterbi decoding algorithm, allowing inference-time refinement without retraining or added model complexity. Our approach improves both fully and semi-supervised TAS models by correcting structural prediction errors while maintaining high efficiency. Code is available at \url{https://github.com/LUNAProject22/CAD}

\end{abstract}
    
\input{sec/1_intro}
\input{sec/2_literature}
\input{sec/3_method}
\input{sec/4_experiments}

\input{sec/5_conclusion}

{\small
\bibliographystyle{unsrt}
\bibliography{main}
}

\end{document}

%% file: sec/1_intro.tex
\section{Introduction}

Temporal action segmentation (TAS) is the task of partitioning an untrimmed video into a sequence of contiguous segments, each labeled with a specific action class. This task is fundamental for understanding long-form human activities in domains such as surveillance, robotics, autonomous driving, human-computer interaction, and instructional video analysis. With the proliferation of publicly available videos and the increasing need for continuous video analysis, TAS has become an essential component in practical video understanding systems.
While existing action recognition models ~\cite{Wu2015WatchnpatchUU,Lin2018TSMTS,Shiota2024EgocentricAR} perform well on short, pre-segmented clips, they fall short in realistic applications where actions appear in succession, vary in duration, and lack explicit boundaries. These challenges are further amplified by the high cost of obtaining dense, frame-level annotations. To mitigate this, weakly supervised approaches that utilize transcripts, which are ordered lists of actions without frame-level alignment, have gained attention for their ability to reduce annotation costs while still enabling effective learning.


\begin{figure}[t!]
\centering
\begin{minipage}{0.38\textwidth}  
    \centering
    \includegraphics[width=\linewidth]{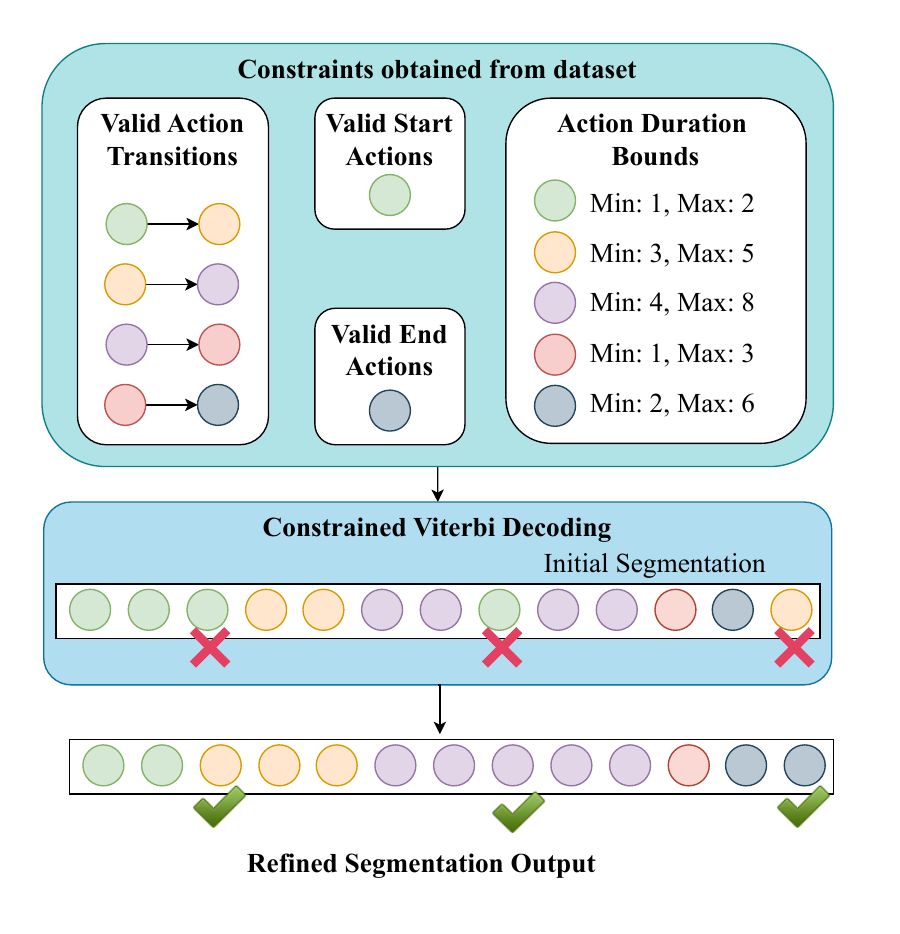}
    \caption{Constraint-aware Viterbi decoding improves TAS by preventing invalid transitions and unstable segment lengths. By enforcing valid start/end actions, allowed transitions, and duration limits, it selects the most probable path that is also structurally consistent, producing temporally coherent and interpretable segmentation.}
    \label{fig:diagram}
\end{minipage}
\hfill
\begin{minipage}{0.58\textwidth}  
    \centering
    \includegraphics[width=\linewidth]{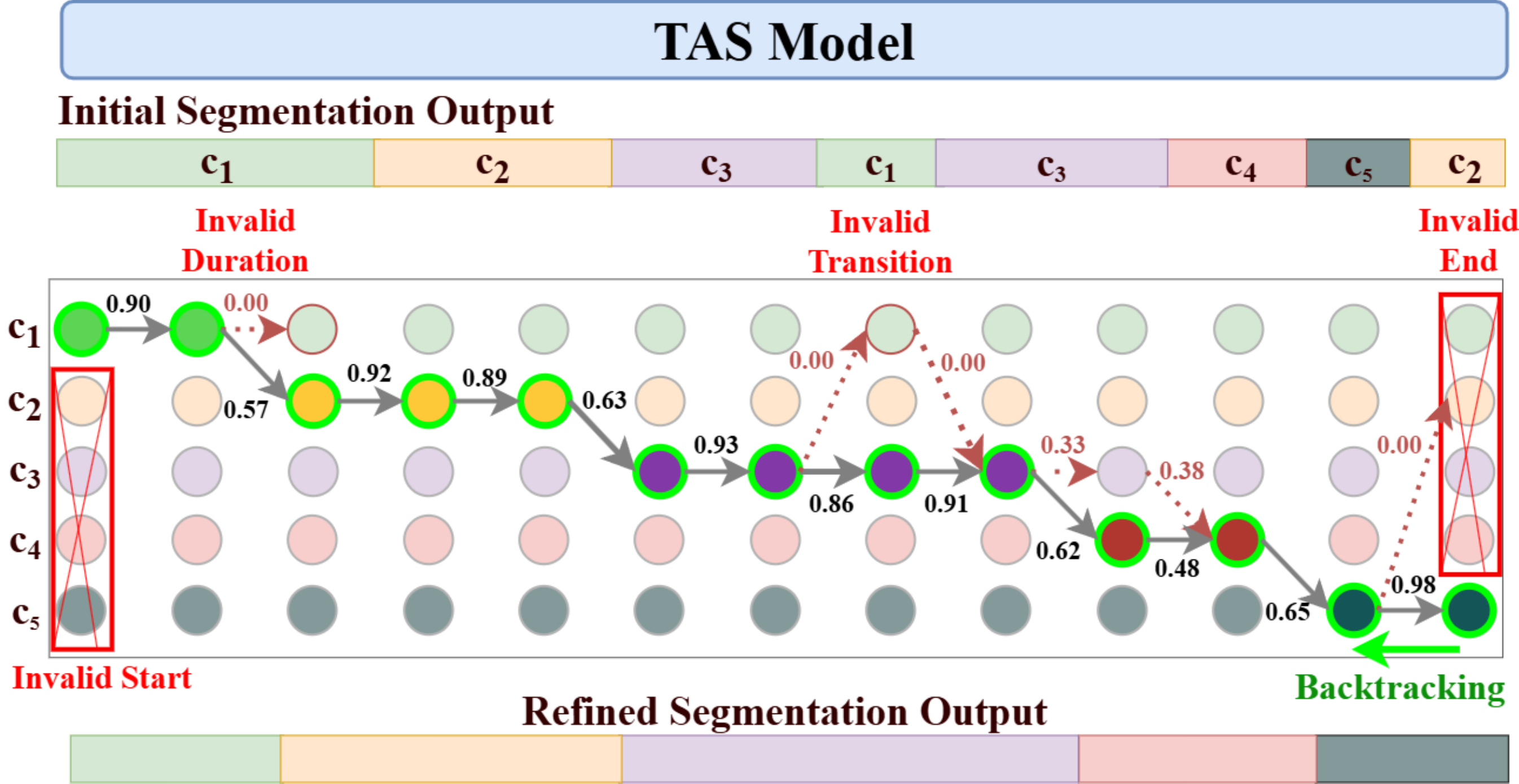}
    \caption{Overview of the proposed constraint-aware temporal action segmentation. Structural constraints (valid starts/ends, action transitions, and duration bounds) are enforced within Viterbi decoding, eliminating invalid transitions and guiding backtracking to the optimal path. Red arrows denote low-probability transitions, black arrows valid transitions, and green arrows the final backtracking flow, yielding consistent and interpretable segmentation. Best viewed in color.}
    \label{fig:overall_pipeline}
\end{minipage}
\end{figure}

Despite advances in temporal modeling, state-of-the-art segmentation methods often suffer from over-segmentation~\cite{Ishikawa2020AlleviatingOE}, poor temporal coherence, and limited generalization. To address this, recent works incorporate structural priors via temporal grammars or hierarchical models. Grammar-based frameworks like KARI~\cite{gong2023activity} encode high-level temporal structures and enforce semantic constraints through rule-based parsing. These priors reduce implausible transitions, improve global consistency, and enhance interpretability, but typically rely on complex recursive grammars and costly parsing algorithms.
We propose a lightweight, constraint-based refinement framework that extracts structural constraints directly from annotated data, including high-confidence temporal transitions, valid start/end actions, and per-class duration priors (Fig.~\ref{fig:diagram}). These weak, statistically derived priors regularize predictions while allowing natural variation, are commonly used in action segmentation~\cite{AbuFarha2019MSTCNMT,li2020set,li2021anchor}, and can be automatically applied to new datasets. By avoiding symbolic grammar induction and recursive parsing, our approach offers a simpler and more efficient solution.
We formulate TAS as a constrained optimization problem over predicted probabilities and solve it with a modified Viterbi algorithm. This algorithm enforces valid transitions, segment durations, and boundary conditions at inference time without changing the TAS model. Unlike approaches requiring extra training or neural Viterbi modules~\cite{Richard2018CVPR}, our method is training-free, low-cost, and suitable for large-scale or real-time use. Experiments on 50Salads and Breakfast show consistent improvements in frame-wise accuracy, segmental edit distance, and F1 scores. In semi-supervised settings (1–5\% labeled data), it refines pseudo-labels to enhance segmentation, while in fully supervised settings, it corrects structural errors without altering model weights. Runtime analysis further shows our method is faster than grammar-based refinement while achieving comparable or better accuracy at lower complexity.

Our contributions are summarized as follows: (1) We propose a constraint-based refinement framework that integrates statistical structural priors into the TAS decoding process, instead of deploy the complexity of grammar-based models.
(2) We design a modified Viterbi algorithm that efficiently incorporates multiple constraints such as valid transitions, durations, and boundary conditions during inference, without retraining the model.
(3) We conduct experiments on standard TAS benchmarks (50Salads and Breakfast) under both semi-supervised and fully supervised settings, showing that our method consistently enhances temporal segmentation quality with minimal overhead.

%% file: sec/2_literature.tex
\section{Related Work}

\paragraph{Temporal Action Segmentation.}
Temporal action segmentation (TAS) aims to divide long, untrimmed videos into a sequence of temporally ordered and semantically labeled action segments. 
Early approaches relied on hand-crafted features and probabilistic models, such as Hidden Markov Models (HMMs) ~\cite{Kuehne16end,Sener2015UnsupervisedSP} and temporal Conditional Random Field  (CRFs)~\cite{Mavroudi2018}. 
Recent progress has been driven by deep learning models, including RNN ~\cite{Richard2017WeaklySA}, diffusion methods ~\cite{liu2023diffusion}, Temporal Convolutional Networks (TCNs) ~\cite{Lea2016TemporalCN}, graph-based methods~\cite{Huang2020ImprovingAS}, and Transformer-based architectures~\cite{chinayiASformer}. These models typically require dense frame-level annotations and focus on improving temporal modeling and long-range dependencies.
Weakly supervised action segmentation methods a solution but their performance comparatively lower than supervised methods~\cite{ng2021weakly}. Some weakly supervised action forecasting methods also has been explored before~\cite{ng2020forecasting} and these methods could benefit from our proposed method.

\paragraph{Semi- and Fully-Supervised Learning}
Temporal action segmentation must capture long-range dependencies while maintaining precise boundaries. Fully supervised methods achieve this by learning temporal coherence from dense frame-level annotations, eliminating handcrafted post-processing. Advanced architectures like MS-TCN~\cite{AbuFarha2019MSTCNMT} refine predictions through stacked dilated temporal convolutions, MS-TCN++~\cite{Li2020MSTCNMT} extends receptive fields with dual dilated layers, and ASFormer~\cite{chinayiASformer} uses hierarchical temporal transformers to capture both global and local context, achieving state-of-the-art performance.
However, dense annotations are costly, motivating weakly- and semi-supervised methods. Transcript-supervised approaches align unordered action transcripts to frames~\cite{Huang2020ImprovingAS}, while semi-supervised methods use limited labeled data and pseudo-labeling to iteratively refine predictions~\cite{singhania2022iterative,10147035,Ding2022LeveragingAA}. These approaches often suffer from over-segmentation and boundary errors, motivating the use of temporal constraints, structural priors, or decoding strategies to improve segmentation quality.

\paragraph{Grammar-Based \& Constraint-Based Modeling}

Early Temporal Action Segmentation (TAS) methods often rely on frame-wise classification followed by smoothing or Conditional Random Fields (CRFs). To improve performance, several works incorporate structural priors or constraints. Grammar-based approaches, for instance, enforce syntactic rules over action sequences to model global ordering. \cite{Kuehne2014TheLO} utilizes context-free grammars for action transitions, while \cite{piergiovanni2020differentiable} integrates them into a differentiable neural framework. Further extensions include adversarially trained generative grammars for activity prediction \cite{piergiovanni2020adversarial} and stochastic grammars to capture hierarchical relationships \cite{qi2017predicting}. More recently, \cite{gong2023activity} introduced a neuro-symbolic framework using probabilistic context-free grammar induction to refine segment boundaries. Similarly, \cite{Pirsiavash2014ParsingVO,Vo2014FromSG} integrate neural sequence modeling with symbolic grammars to align predictions with temporal constraints.
Beyond grammars, logic-based modeling provides a model-agnostic alternative. For example, Differentiable Temporal Logic (DTL) \cite{NEURIPS2022} embeds constraints into deep networks to enhance coherence without architectural changes. Motivated by these structured priors, we propose a simpler constraint-guided decoding strategy. Our approach explicitly encodes structural constraints during inference, offering a lightweight alternative to improve temporal consistency and mitigate over-segmentation errors.
Our method also falls under methods that explore logical constraints for various tasks such as procedural question answering~\cite{nguyen2025neuro} and vision action learning~\cite{choi2025nesyc}. 

%% file: sec/3_method.tex
\section{Methodology}

We propose a lightweight and interpretable framework for refining the output of a temporal action segmentation (TAS) model by incorporating domain-specific structural constraints. 
Our approach enhances temporal coherence and eliminates implausible predictions without modifying the underlying TAS backbone. 
At first, we extract structural constraints such as transition confidence and other domain-specific constraints such as valid start/end actions and action duration bounds. 
Then, we perform constrained inference using a modified Viterbi decoding algorithm that enforces all constraints while optimizing the predicted action sequence for optimal reassignment of category labels.
The overall pipeline of the proposed framework is illustrated in Fig.~\ref{fig:overall_pipeline}.


\subsection{Extracting Structural Constraints from Activity Videos}

Structural constraints define the structure of activities in terms of its constituent actions. 
One such structural constraint that we consider are frequently occurring action transitions.
Specifically, for each consecutive pair of actions \( (A \rightarrow B) \) in the video, we compute \textbf{transition confidence} as:

\begin{equation}
\text{Conf}(A \rightarrow B) = \frac{\text{Count}(A \rightarrow B)}{\text{Count}(A)}.
\end{equation}

All observed transitions are included in the set of valid transitions $\mathcal{T}$ with their empirical confidences directly computed to ensure reliability.
This results in a sparse, interpretable transition graph that reflects temporal patterns of activities in the dataset. These transitions serve as the foundation for constraining action sequences during inference.

In addition to transition confidence, we consider several other structural priors that reinforce action sequence plausibility and enforce semantic structure:

\begin{itemize}
    \item \textbf{Valid Start Actions:} A set of action classes \( \mathcal{S}\) observed as the starting action of activities in the dataset videos.
    \item \textbf{Valid End Actions:} A set of action classes \( \mathcal{E}\) observed as the last action of activities in the dataset videos.
    \item \textbf{Normalized Action Duration Bounds:} For each action class \( c \) (e.g. \textit{take\_cup, pour\_milk}), we define a valid duration interval \([d^{\min}_c, d^{\max}_c]\), normalized relative to video length of the entire activity.
\end{itemize}

The constraints described such as transition confidence, valid start and end actions, and normalized duration bounds are all directly computed from the annotated videos in the dataset. 
Each of them relies on simple counting statistics or empirical observations, without requiring complex modeling or external supervision. 
Also, these constraints are based on data-driven statistics rather than dataset-specific heuristics, they remain interpretable and transparent to practitioners. 
Each constraint has a clear semantic meaning such as “action A usually follows action B” or “action C typically lasts between 5–10\% of the video length.” This interpretability makes it easy to inspect, debug, or even manually adjust the constraints if needed.
With these constraints we are able to cover a large number of complex sequence dynamics without using complex grammar or logic induction methods.
Finally, these constraints generalize well across different datasets or domains because they are derived from the structure of the data itself. 
As long as a dataset contains annotated sequences of actions, the same procedure can be used to extract structural priors.

\subsection{Constraint-Aware Viterbi Decoding}
We integrate the extracted structural constraints described above to correct prediction errors in action segmentation by introducing a constraint-aware variant of the Viterbi algorithm. 
The standard Viterbi algorithm is a dynamic programming approach that finds the most probable sequence of actions given a sequence of frame-wise action scores or probabilities.
At each time step, it updates the best score for each state by considering the highest cumulative score from all possible previous states, combined with the corresponding transition and observation probabilities. 
In addition, the algorithm maintains backpointers to record the optimal preceding state for each current state, enabling reconstruction of the highest-scoring path after processing all frames.




We extend the classical Viterbi algorithm to find the most probable sequence of action labels for an input video, given the frame-wise action probabilities from a TAS model and a set of structural constraints. 
Let \( P \in \mathbb{R}^{T \times C} \) denote the frame-wise action probability matrix produced by a TAS backbone, where \( T \) is the number of frames and \( C \) is the number of action classes.  
We define a set of structural constraints \( R \), which includes valid start and end actions, permissible action transitions (with their empirical confidences), and normalized duration bounds for each action class. 
The objective is to find the optimal label sequence that maximizes the joint probability of action classes and valid transitions while satisfying all constraints:

\[
Y^* = \arg\max_{Y \in \mathcal{Y}_R} \left( \sum_{t=1}^{T} \log P_{t, y_t} + \sum_{t=2}^{T} \log \text{Conf}(y_{t-1} \rightarrow y_t) \right)
\]

\noindent \text{subject to:}
\begin{align*}
& y_1 \in \mathcal{S} \quad \text{(valid start actions)} \\
& y_T \in \mathcal{E} \quad \text{(valid end actions)} \\
& (y_{t-1} \rightarrow y_t) \in \mathcal{T}, \quad \forall t \in [2, T] \quad \text{(valid transitions)} \\
& d^{\min}_c \leq \frac{t_e - t_s + 1}{T} \leq d^{\max}_c, \quad \forall \text{ segment } [t_s, t_e] \text{ with label } c
\end{align*}

\noindent Here, \( \mathcal{Y}_R \) denotes the set of all label sequences satisfying the structural constraints \( R \).



The decoding consists of three main stages: initialization, forward pass, and backtracking. We define the following matrices:

\begin{itemize}
    \item \( V \in \mathbb{R}^{T \times C} \): stores the maximum log-probability of a sequence ending with action class \( c \) at time \( t \).
    \item \( D \in \mathbb{R}^{T \times C} \): tracks the duration of the current action segment up to time \( t \).
    \item \( B \in \mathbb{N}^{T \times C} \): stores backpointers used for recovering the optimal sequence.
\end{itemize}

\paragraph{Initialization.} At time step \( t=0 \), only the action classes belonging to the valid start set \( \mathcal{S} \) are assigned non-zero scores based on their emission probability. All other classes are set to \( -\infty \), effectively disallowing them as starting actions:

\begin{equation}
V[0, c] = 
\begin{cases}
\log P[0, c], & \text{if } c \in \mathcal{S} \\
-\infty, & \text{otherwise}
\end{cases}
\end{equation}

The duration tracker is initialized such that start actions begin with duration 1:

\begin{equation}
D[0, c] = 
\begin{cases}
1, & \text{if } c \in \mathcal{S} \\
0, & \text{otherwise}
\end{cases}
\end{equation}

\paragraph{Forward Pass.}  
For each time step \( t = 1 \) to \( T-1 \), and each action class \( c \), we compute the best possible transition into \( c \) from any previous class \( c' \), constrained by the transition set \( \mathcal{T} \) and duration bounds.

\textbf{Self-transition case} (\( c = c' \)):  
If the current action is the same as the previous, it is treated as a continuation of the same segment. We only allow self-transitions if the normalized duration is within the maximum allowed i.e. \( \frac{D[t-1, c]}{T} < d^{\max}_c \) then
\( 
V[t, c] = V[t-1, c] + \log P[t, c], D[t, c] = D[t-1, c] + 1, 
B[t, c] = c. \)

\textbf{Action-transition case} (\( c \ne c' \)):  
We consider a transition only if the current segment in \( c' \) has reached at least its minimum duration \( \frac{D[t-1, c']}{T} \geq d^{\min}_{c'} \) and \( (c' \rightarrow c) \in \mathcal{T} \) then:
\begin{align}
\text{score} &= V[t-1, c'] + \log \text{Conf}(c' \rightarrow c) + \log P[t, c] 
\end{align}
If score \( > V[t, c] \)  then update: \( V[t, c] = \text{score}, B[t, c] = c', 
D[t, c] = 1. \)

This process is repeated across all valid \( c' \) for each \( c \), ensuring only structurally consistent transitions and durations are considered.

\paragraph{Termination.}  
At the final time step \( T-1 \), we identify the best-scoring action class from the valid end set \( \mathcal{E} \), again ensuring that duration constraints are satisfied:

\begin{equation}
c^* = \arg\max_{c \in \mathcal{E}, \; \frac{D[T-1, c]}{T} \leq d^{\max}_c} V[T-1, c]
\end{equation}

\paragraph{Backtracking.}  
The optimal sequence \( \hat{Y}_{1:T} \) is then recovered by tracing the backpointers in reverse, starting from \( c^* \) and using the matrix \( B \):

\begin{equation}
\hat{Y}[t] = B[t+1, \hat{Y}[t+1]] \quad \text{for } t = T-2, T-3, \dots, 0
\end{equation}

As we apply the constrained decoding on the predicted logits \( P \) of a TAS model our framework is compatible with any segmentation backbone as we show in Table \ref{tab:ablation_on_fact}.
Our method offers a simple yet powerful decoding strategy that injects structural knowledge into TAS predictions using data-derived constraints. 


%% file: sec/4_experiments.tex
\section{Experiments}

We evaluate the effectiveness of our proposed rule-based refinement framework under both \textit{semi-supervised} and \textit{fully supervised} settings. 
In both cases, our method is applied as a post-processing step on top of existing temporal action segmentation (TAS) models, demonstrating its versatility and capacity to improve predictions without requiring model modification or retraining (except for semi-supervised setting). 
This section details the datasets used in our experiments, followed by implementation details under the two supervision settings.

\subsection{Datasets and evaluation metrics}

We conduct experiments on two widely-used benchmark datasets for temporal action segmentation:
\textbf{Breakfast}~\cite{kuehne2014language},
    a large-scale dataset comprising 1,712 video recordings, documents 52 participants as they carry out 10 types of breakfast-related tasks. These tasks are broken down into 48 fine-grained actions and were filmed in 18 separate kitchen settings.
\textbf{50Salads}~\cite{stein2013combining}, contains 50 egocentric videos of salad preparation activity performed by 25 different subjects. 
There are total 19 fine-grained actions.

For evaluation, we adopt the standard metrics: F1 score at multiple IoU thresholds (F1@10, F1@25, F1@50), segmental edit distance (Edit), and frame-wise accuracy (Acc).
These metrics collectively capture the precision, temporal coherence, and alignment quality of the predicted segmentation.

\subsection{Semi-supervised Setting}





In the semi-supervised setting, we evaluate our method on top of the temporal action segmentation model proposed by ~\cite{10147035}, which employs an iterative contrast-classify (ICC) framework ~\cite{singhania2022iterative}. 
This model alternates between unsupervised representation learning and supervised classification using pseudo-labels, and has demonstrated strong performance under limited supervision.

\begin{figure}
    \centering
    \includegraphics[width=0.90\textwidth]{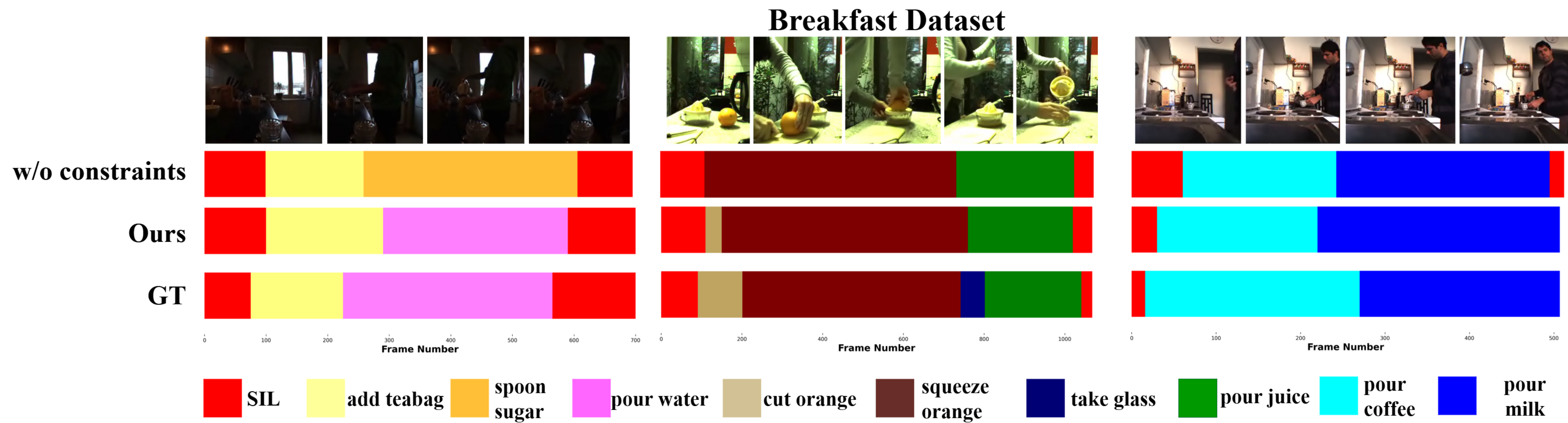}
    
     \includegraphics[width=0.90\textwidth]{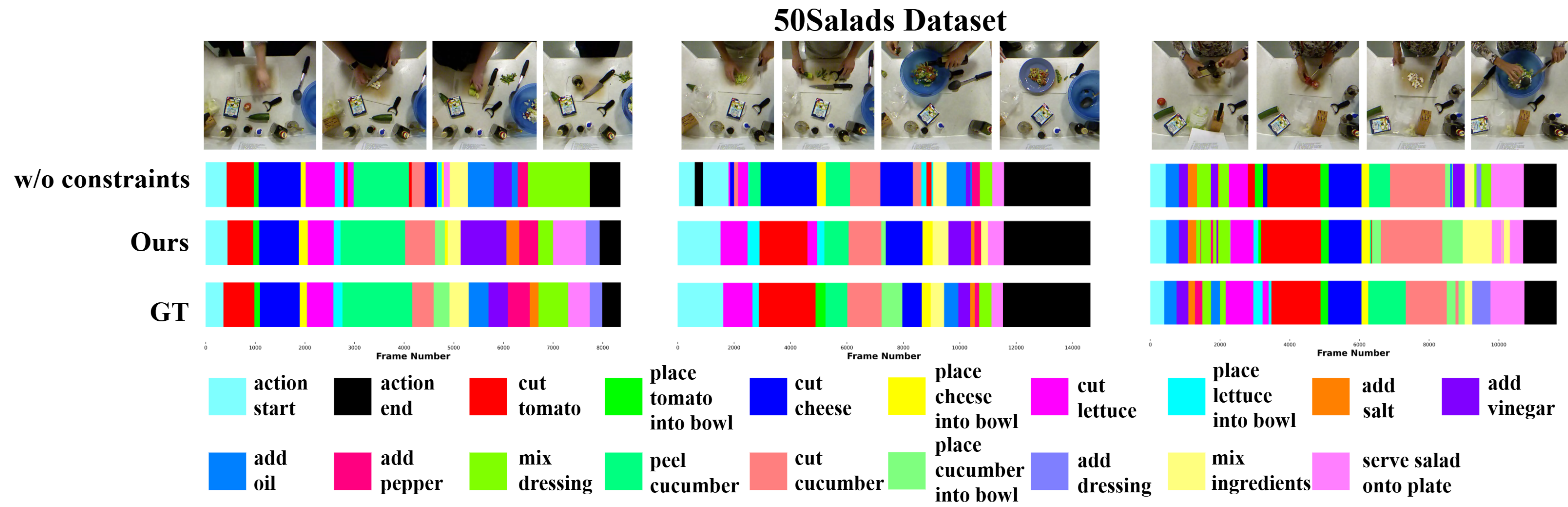}
    \caption{Qualitative examples from the semi-supervised experiments demonstrate that our approach produces segmentation outputs with better alignment to the ground truth (GT), after training with constraint-based refined pseudo-labels during iterative learning. The top row shows the output from the TAS model without any constraint-aware refinement, which contains invalid action segments. Best viewed in color.}
    \label{fig:qualitative_diagrams}
\end{figure}

Our refinement approach is applied as a post-processing step to enhance the pseudo-labels generated in each iteration of the ICC framework. 
We conduct experiments on the 50Salads and Breakfast datasets, both of which are standard benchmarks in temporal action segmentation. 
Each dataset provides diverse, untrimmed videos annotated with fine-grained action sequences, but no frame-level annotations are used for generating pseudo-labels.
In Table \ref{tab:overall_table}, we show that incorporating constraints-based refinement consistently improves performance across all metrics compared to the baseline ICC model without constraints. 
Notably, we observe gains in Edit and F1 scores, indicating improved temporal coherence and structural plausibility of the predicted sequences.
These findings suggest that simple, interpretable structural constraints can provide valuable regularization in semi-supervised learning by improving the quality of pseudo-labels for unlabeled data.
Our method converges faster in semi-supervised settings because the constraints provide prior knowledge that guides pseudo-label generation, yielding higher-quality labels at each iteration and thus accelerating learning. Without constraints, the model relies only on its own predictions, requiring more iterations to reach optimal performance.
Fig.~\ref{fig:qualitative_diagrams} shows the qualitative examples of the output from the TAS model.

\subsection{Fully-supervised Setting}

To demonstrate that our method also complements strong supervised models, we apply it in a fully supervised setting using existing TAS models \textsc{ASFormer}~\cite{chinayiASformer}.
We evaluate our method on the 50Salads and Breakfast datasets
Although the model are trained with full supervision and achieve state-of-the-art accuracy, it still exhibit common issues such as over-segmentation, where a single action is split into multiple short fragments, and temporally implausible transitions between semantically unrelated actions as shown in the qualitative example.
By applying our constrained-based decoding as a refinement step, we correct these artifacts and produce more coherent action sequences. 
As shown in Table \ref{tab:overall_table}, our method particularly improves Edit and F1 metrics, as it directly targets structural errors and temporal misalignment rather than isolated frame-level mistakes.
Importantly, we emphasize that our method is \textit{model-agnostic} and \textit{training-free} during inference. It can be seamlessly integrated into any existing TAS pipeline to enhance robustness and consistency.

To assess the benefits of our rule-based refinement, we compare our approach with the grammar-based method KARI ~\cite{gong2023activity}, which integrates context-free grammar induction and parsing for temporal action segmentation. 
They induce a probabilistic grammar by identifying key actions that frequently appear in the annotated sequences and grouping actions around them. 
Their method constructs complex production rules, including recursive OR-rules and AND-rule chains, to capture valid temporal structures. 
Finally, parser design, thus offering a lightweight yet effective solution.
In contrast, our method leverages simple constraints directly extract from the dataset sequences without heuristic key action selection or recursive constructions. 
This results in a more interpretable and scalable constraints set that covers essential temporal constraints with significantly reduced complexity.

We evaluate both KARI and our constraint-aware decoding using standard segmentation metrics: frame-wise accuracy (Acc), segmental edit distance (Edit), and F1 scores at IoU thresholds 10\%, 25\%, and 50\%. 
Results indicate that our approach consistently achieves better segmentation quality than the grammar-based refinement.
\input{tables/semi_fully_sup_result_table}

\subsection{Complexity and Runtime Comparison}
Several grammar-based approaches have been proposed for modeling activity structure in videos. Differentiable grammars~\cite{piergiovanni2020differentiable} and adversarial generative grammars~\cite{piergiovanni2020adversarial} integrate grammar learning into deep networks, while stochastic grammars~\cite{qi2017predicting} capture hierarchical and temporal relations. The KARI method~\cite{gong2023activity} uses a Breadth-first Earley Parser (BEP) with heuristics like queue pruning. However, these grammar-based techniques require complex parsing and have worst-case complexity 
$(\mathcal{O}(T^3)$, where $T$ is the number of frames.

\input{tables/infer_complex_table}

In contrast, our method avoids grammar parsing by applying structural constraints (valid transitions, start/end actions, duration bounds) directly in a modified Viterbi algorithm. This yields efficient decoding with complexity 
$\mathcal{O}(CT)$ ($C$ is the number of action classes), effectively linear in 
$T$, and easily integrates with any TAS model without grammar induction or retraining.
We evaluate efficiency by comparing inference time with the grammar-based method~\cite{gong2023activity} on 20 Breakfast videos using the same TAS backbone and an NVIDIA A5000 GPU. As shown in Table~\ref{infer_table} and Fig.~\ref{fig:comparison_runtime}, our approach is significantly faster while maintaining comparable or better segmentation accuracy.

\subsection{Ablation}

\textbf{Effectiveness of Constraint-Based Decoding.}
To evaluate the impact of additional structured constraint decoding, we conduct an ablation study comparing our constraint-based Viterbi decoding with the classical Viterbi algorithm in the semi-supervised setting. 
The classical Viterbi relies on TAS action logits for each time step and transition confidences for sequence prediction. 
Our results show that using the classical Viterbi algorithm lags behind our constrained Viterbi decoding approach.
This observation highlights the importance of structural constraints to effectively guide the decoding process toward more plausible temporal structures and prevent degenerate solutions, especially in low-supervision settings where the model may otherwise overfit noisy or ambiguous patterns. 
Incorporation of structural constraints during decoding plays a critical role in enhancing prediction quality.

\input{tables/ablation_classical_viterbi}

\begin{figure}
    \centering
    \includegraphics[width=\textwidth]{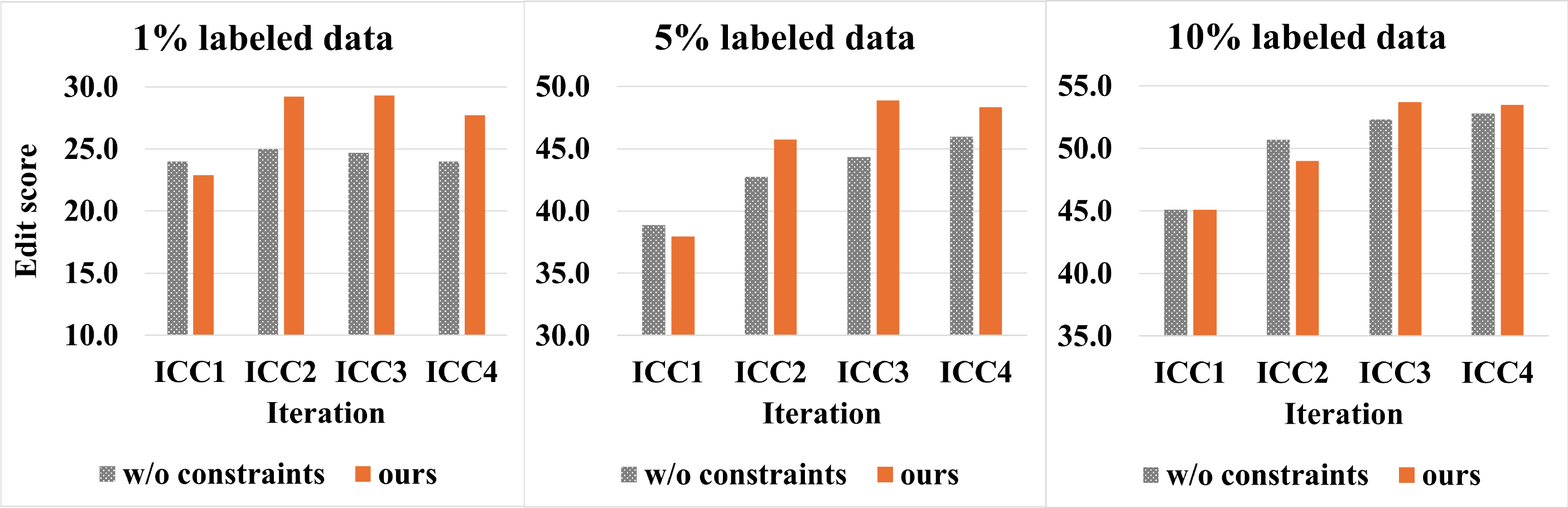}
    \caption{Comparison of edit score improvements across ICC iterations in a semi-supervised setting using different amounts of labeled data (1\%, 5\%, and 10\%). Our constraint-based method consistently outperforms the baseline (w/o constraints) across all iterations and data proportions. Best viewed in color.}
    \label{fig:ablation_1_10_data_50salads}
\end{figure}

\noindent
\textbf{Effectiveness of Constraint-Based Learning in Semi-supervised setting.}
To demonstrate that the proposed constraint-based method is highly effective in semi-supervised settings, we use 1\%, 5\% and 10\% labeled data on the 50Salads dataset to perform action segmentation on the unlabelled data.
As shown in Fig.~\ref{fig:ablation_1_10_data_50salads} which illustrates only the edit score across ICC iterations,
even with only 1\% of labeled data (equivalent to a single annotated video in 50 Salads dataset), the model augmented with structural constraints significantly outperforms the baseline ICC \cite{singhania2022iterative} method that without any pseudo-label refinement throughout the semi-supervised learning iterations.
Although this figure focuses solely on the edit score metric, we provide a comprehensive comparison including additional performance metrics (e.g., F1@{10, 25, 50}) in the supplementary material for completeness.
This strong performance continues as the amount of labeled data increases to 5\% and 10\%, consistently surpassing the baseline and validating the robustness of our approach. 
These findings highlight the key role of constraints in guiding the learning process when annotated data is scarce. 
By leveraging structural and temporal priors, the model can generalize meaningful patterns from very limited supervision. 
This suggests that constraint-based decoding is an effective solution where manual annotation is expensive.
\newline
\textbf{Compatibility with different TAS Backbones.}
To evaluate the generalizability of our method, we further investigate its compatibility with alternative temporal action segmentation (TAS) backbones. 
Specifically, we integrate our constraint-based decoding into the FACT model and evaluate its performance in the fully-supervised setting on the Breakfast dataset. 
As shown in Table~\ref{tab:ablation_on_fact}, our method achieves comparable performance across multiple metrics compared to the baseline model trained without any constraints-based optimization. 
These results demonstrate that our approach can be seamlessly integrated with different model backbones, making it a versatile addition to existing TAS frameworks without requiring major architectural changes.

\input{tables/ablation_fact_backbone}

\noindent
\textbf{Ablation on constraints.}
We performed per-component ablations to quantify the contribution of valid start/end actions, permitted transitions, and per-class duration bounds (Table~\ref{tab:rebuttal_ablations_asf_table}). Removing duration constraints produces the largest drop in F1, while removing transition constraints primarily reduces edit and accuracy, indicating complementary roles: duration bounds stabilize segment lengths, and transition rules improve temporal consistency. 
Varying the relative weights of the transition and duration terms in the decoding objective yielded negligible differences from the balanced setting, demonstrating robustness to weighting choices and that the hard constraints themselves drive the improvements. We also implemented a soft-constraint variant (softmax-based relaxation with penalties for violations); because invalid transitions and durations retain small nonzero probabilities in this formulation, it consistently underperforms hard-constraint design, confirming the effectiveness of enforcing feasibility. We explored richer constraints such as multi-parent dependencies (e.g., $a_1, a_2, a_3 \rightarrow a_4$) but observed no gains; we therefore focus on constraints that are reliably extractable across datasets, leaving hierarchical or context-specific extensions for future work.

%% file: tables/semi_fully_sup_result_table.tex
\begin{table}
\centering
\small
\begin{tabular}{c|ccccc|ccccc}
\hline
& \multicolumn{5}{c|}{Breakfast} & \multicolumn{5}{c}{50Salads} \\
\cline{2-11}
\multirow{-2}{*}{Method} & \multicolumn{3}{c}{$F1@\{10,25,50\}$} & Edit & Acc. & \multicolumn{3}{c}{$F1@\{10,25,50\}$} & Edit & Acc. \\
\hline

\multicolumn{11}{c}{Semi-supervised setting.} \\

\hline

w/o constraints
& 57.1 & 51.2 & 34.6 & 54.6 & \textbf{64.2}
& 51.8 & 47.7 & 37.0 & 45.9 & 62.3\\

Ours
& \textbf{58.5} & \textbf{52.4} & \textbf{36.0} & \textbf{56.4} & 63.9
& \textbf{55.8} & \textbf{51.4} & \textbf{38.5} & \textbf{48.9} & \textbf{63.4} \\

\hline

\multicolumn{11}{c}{Fully-supervised setting.} \\

\hline

w/o constraints
& 74.1 & 68.7 & 55.5 & 72.8 & 72.4
& 83.4 & 80.8 & 74.6 & 75.7 & \textbf{85.0} \\

KARI* \cite{gong2023activity}
& 77.3 & 71.6 & 57.0 & 78.3 & 74.5
& 83.8 & 81.8 & 74.7 & 76.4 & 83.3 \\

Ours 
& \textbf{78.8} & \textbf{73.1} & \textbf{58.0} & \textbf{77.9} & \textbf{74.6} 
& \textbf{84.8} & \textbf{83.1} & \textbf{76.0} & \textbf{78.6} & 82.9 \\

\hline
\end{tabular}
\caption{{Performance results on both semi-supervised (5\% of labeled data) and fully supervised settings. As we compare our method with the counterparts under different settings, we follow the work ~\cite{singhania2022iterative} in semi-supervised setting and using the C2F-TCN ~\cite{10147035} model as TAS backbone, whereas we follow ~\cite{gong2023activity} for fully-supervised setting and using ASFormer ~\cite{chinayi_ASformer} as backbone. For the KARI* method, we reproduced the segmentation results using the given induced grammar rules.}}
\label{tab:overall_table}
\end{table}













                    


%% file: tables/infer_complex_table.tex
\begin{figure}[t]
\centering
\begin{minipage}{0.48\textwidth}  
    \scriptsize
    \begin{tabular}{p{3.8cm}c}
        \toprule
        \textbf{Method} & \textbf{Time (s)} \\
        \midrule
        KARI~\cite{gong2023activity} (grammar + BEP parser) & 148.9 \\
        \textbf{Ours} (constrained Viterbi) & \textbf{2.81} \\
        \bottomrule
    \end{tabular}
    \captionof{table}{Average inference time comparison using ASFormer for 20 sample videos from Breakfast under fully supervised setting.}
    \label{infer_table}
\end{minipage}
\hfill
\begin{minipage}{0.48\textwidth}  
    \includegraphics[width=\linewidth]{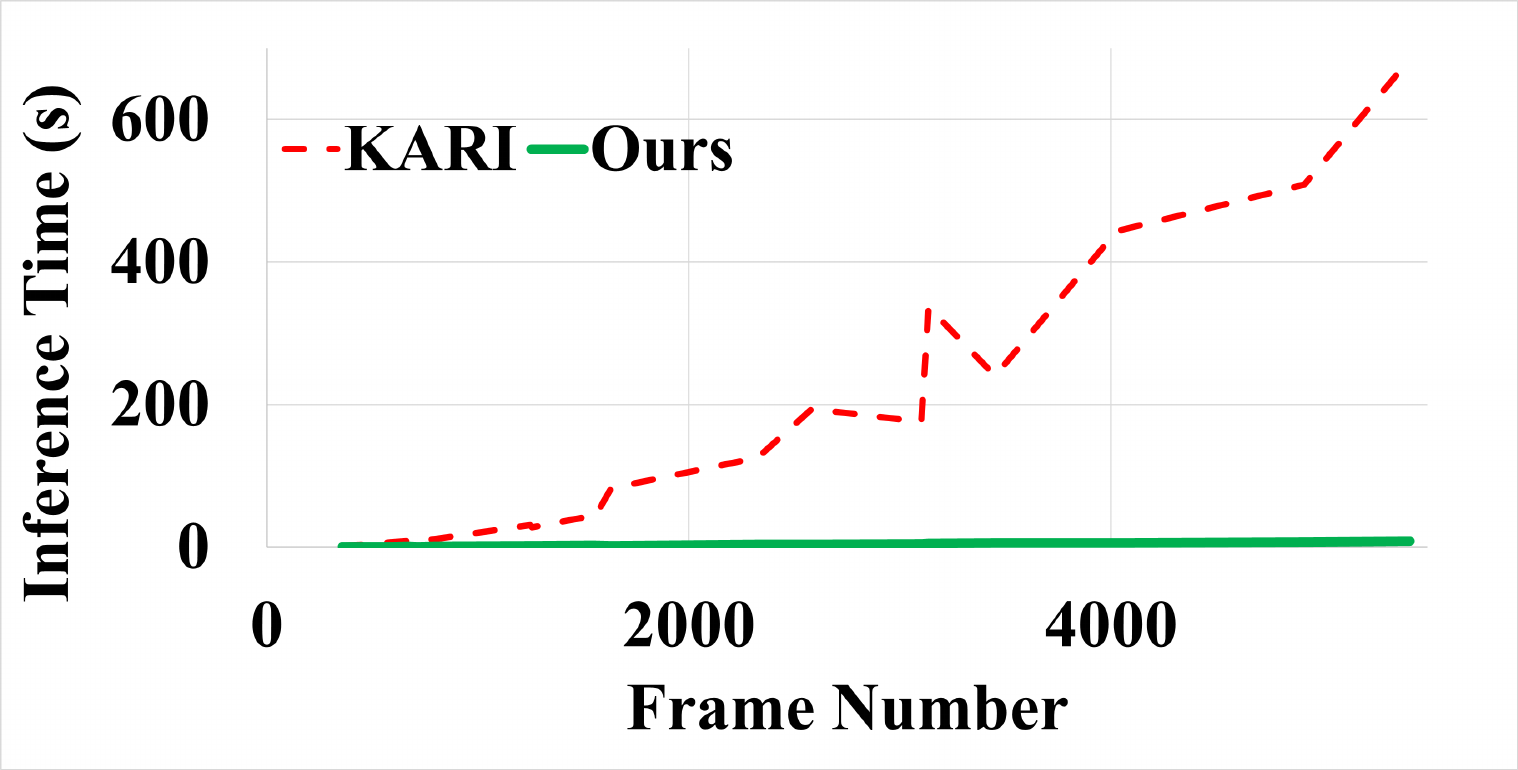}
    \caption{Inference time scaling with video length.}
    \label{fig:comparison_runtime}
\end{minipage}
\end{figure}

%% file: tables/ablation_classical_viterbi.tex
\setlength{\tabcolsep}{3pt} 
\begin{table}
\scriptsize
\centering
\begin{tabular}{c|ccccc}
\hline
& \multicolumn{5}{c}{50salads} \\
\cline{2-6}
Method & \multicolumn{3}{c}{$F1@\{10,25,50\}$} & Edit & Acc. \\

\hline
                 
Classical Viterbi 
& 45.5 & 40.2 & 30.1 & 39.2 & 60.1 \\

Ours 
& \textbf{55.8} & \textbf{51.4} & \textbf{38.5} & \textbf{48.9} & \textbf{63.4} \\

\hline
\end{tabular}
\caption{{Ablation on Effectiveness of Constraint-Based Decoding}}
\end{table}



                    



%% file: tables/ablation_fact_backbone.tex
\setlength{\tabcolsep}{2pt} 
\begin{table}[b!]
\centering
\begin{minipage}{0.48\textwidth}  
\centering
\scriptsize
\begin{tabular}{c|c|ccccc}
\hline
& & \multicolumn{5}{c}{Breakfast} \\
\cline{3-7}
\multirow{-2}{*}{Backbone} & \multirow{-2}{*}{Method} & \multicolumn{3}{c}{$F1@\{10,25,50\}$} & Edit & Acc. \\
\hline
& baseline & 74.1 & 68.7 & 55.5 & 72.8 & 72.4 \\
& KARI \cite{gong2023activity} & 77.3 & 71.6 & 57.0 & \textbf{78.3} & 74.5 \\
\multirow{-3}{*}{ASFormer} & Ours & \textbf{78.8} & \textbf{73.1} & \textbf{58.0} & 77.9 & \textbf{74.6} \\
\hline
& baseline & 81.3 & 76.4 & \textbf{66.2} & 79.7 & 75.2 \\
& KARI \cite{gong2023activity} & 81.5 & 76.4 & 65.6 & \textbf{82.5} & 77.1 \\
\multirow{-3}{*}{FACT} & Ours & \textbf{81.8} & \textbf{76.5} & 64.8 & 81.4 & \textbf{77.7} \\
\hline
\end{tabular}
\caption{Fully-supervised performance on the Breakfast dataset using FACT backbone. Constraint-based refinement improves multiple metrics.}
\label{tab:ablation_on_fact}
\end{minipage}
\hfill
\begin{minipage}{0.38\textwidth}  
\centering
\scriptsize
\begin{tabular}{l|ccccc}
\toprule
& \multicolumn{5}{c}{Breakfast} \\
\cline{2-6}
\multirow{-2}{*}{Const.} & \multicolumn{3}{c}{$F1@\{10,25,50\}$} & Edit & Acc.\\
\hline
w/o any & 74.1 & 68.7 & 55.5 & 72.8 & 72.4 \\
w/o start-end & 77.9 & 72.4 & 57.8 & 76.0 & 73.1 \\
w/o transition & 75.1 & 70.0 & 56.3 & 73.8 & 72.3 \\
w/o duration & 75.0 & 69.4 & 54.9 & 75.5 & 74.5  \\
0.3 tra. + 0.7 dur. & 78.8 & 73.1 & 58.0 & 78.8 & 74.6\\
0.7 tra. + 0.3 dur. & \textbf{79.0} & \textbf{73.2} & \textbf{58.2} & \textbf{78.9} & \textbf{74.7}\\	
Soft & 75.6 & 70.5 & 56.8 & 74.6 & 72.4\\
Hard & 78.8 & 73.1 & 58.0 & 77.9 & 74.6 \\
\bottomrule
\end{tabular}
\caption{Ablation on constraints: effect of each component, weighted and soft constraints on Breakfast dataset.}
\label{tab:rebuttal_ablations_asf_table}
\end{minipage}
\end{table}

%% file: sec/5_conclusion.tex
\section{Conclusion \& Discussion}

We introduced a constraint-based refinement method for temporal action segmentation that leverages structural priors extracted from training videos. By extending the classical Viterbi algorithm, our approach incorporates interpretable constraints such as valid start/end actions, permissible transitions, and normalized duration bounds, enabling accurate and semantically coherent segmentation. It achieves efficient inference through a dynamic programming formulation with linear-time complexity. 
Empirical results demonstrate that our refinement method achieves comparable or improved segmentation performance while significantly reducing inference time. Our work highlights the potential of combining lightweight structural constraints with probabilistic decoding, opening up new avenues for fast and interpretable action understanding in videos.
The main limitation is the effectiveness depends on the quality and coverage of structural constraints extracted from annotated training videos. Because valid transitions, valid start/end actions, and duration bounds are estimated from observed sequences, the method may struggle with rare, unseen, or highly variable action orders at test time. The hard-constraint formulation is efficient and produces strong improvements, but it can also become overly restrictive when the training data is sparse, noisy, or incomplete. 
\newline
\textbf{Acknowledgment} This research is supported by the National Research Foundation, Singapore, under its NRF Fellowship (Award\# NRF-NRFF14-2022-0001).